\documentclass{article}
\usepackage[accepted]{eiml_icml2026}
\AtBeginDocument{\setcitestyle{numbers}}

\usepackage{amsmath,amssymb,amsthm,mathtools}
\usepackage{graphicx,booktabs,tabularx,array,caption,float}
\usepackage[table]{xcolor}
\usepackage[most]{tcolorbox}
\tcbset{colback=white, arc=3pt, boxrule=0.6pt, left=6pt, right=6pt, top=5pt, bottom=5pt}
\newcolumntype{C}{>{\centering\arraybackslash}X}
\usepackage[colorlinks,citecolor=blue,linkcolor=red,urlcolor=magenta]{hyperref}

\icmltitlerunning{ReflectiChain: Epistemic Grounding via Generative World Models}

\begin{document}

\twocolumn[
\icmltitle{ReflectiChain: Epistemic Grounding in LLM-Driven\\ World Models for Supply Chain Resilience}

\begin{icmlauthorlist}
\icmlauthor{Jia Luo}{hust}
\end{icmlauthorlist}

\icmlaffiliation{hust}{School of Foreign Languages,\\ Huazhong University of Science and Technology, Wuhan 430074, China}
\icmlcorrespondingauthor{Jia Luo}{u202317016@hust.edu.cn}
\icmlkeywords{Epistemic Intelligence, Supply Chain Resilience,\\ Generative World Models, LLM Agents, Test-Time Adaptation}
\vskip 0.2in
]

\printAffiliationsAndNotice{}

\begin{abstract}
AI agents in supply chains face a fundamental epistemic gap: large language models (LLMs) interpret policies but lack physical grounding, while reinforcement learning (RL) optimizes flows but is semantically blind to unstructured constraints. We introduce \textsc{ReflectiChain}, bridging this gap through a Generative Supply Chain World Model (SC-WM)---encoding heterogeneous supply networks into a 6-dim graph-latent space with physical conservation---and Double-Loop Learning that separates epistemic uncertainty (KL-trust-region-bounded policy adaptation) from aleatoric uncertainty (stochastic latent rollouts). On Semi-Sim, a 10-node semiconductor benchmark with SIR risk propagation, 6 perturbation types, and 10 policy constraint templates, \textsc{ReflectiChain} improves Rationale Consistency Score by {\color{blue}33.0\%} ($p < 0.0001$, $d = 2.78$), maintains 82.3\% operability under adversarial shocks, and exhibits anti-fragile behavior (+40.2\% gain under moderate pressure). We identify three operational epistemic mechanisms---uncertainty separation, knowledge-boundary detection, and empirical Bayesian policy updating---and discuss five limitation categories.
\end{abstract}

\begin{figure*}[t]
    \centering
    \includegraphics[width=0.85\textwidth]{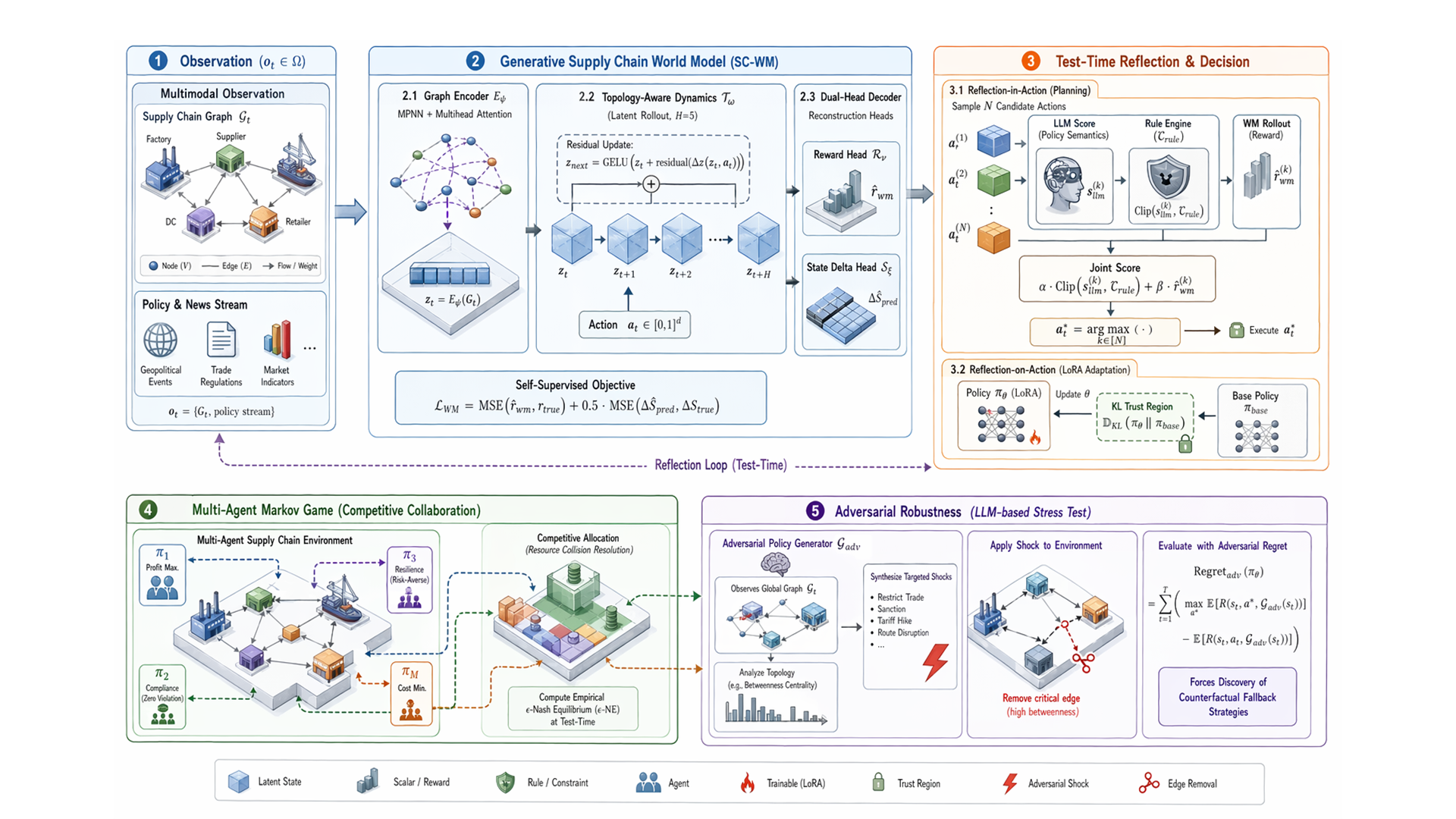}
    \caption{\textbf{ReflectiChain architecture.} (Left) SC-WM: Graph Encoder $\to$ Latent $z_t$ $\to$ Multi-step Rollout $\to$ Dual-Head Decoder. (Right) Double-Loop: Reflection-in-Action ($\mathcal{C}_{rule}$-bounded candidate scoring) + Reflection-on-Action (KL-trust-region LoRA updates).}
    \label{fig:framework}
\end{figure*}

\section{Introduction}

Modern semiconductor supply chains exemplify a critical \textbf{epistemic grounding problem}: when geopolitical policies arrive as unstructured natural-language text, an AI agent must jointly reason about \textit{what constraints mean} and \textit{what actions are physically possible}---bridging semantic and physical knowledge whose representations are fundamentally misaligned. This epistemic gap causes complementary failure modes: \textbf{RL} is \textit{semantically blind}---policies as text never enter its state representation; \textbf{LLMs} suffer \textit{grounding gaps}---they prescribe semantically plausible but physically infeasible actions. Neither system can represent the boundary of its own knowledge.

\textbf{Illustration.} When the CHIPS Act's ``guardrail clause'' prohibits entities receiving U.S. subsidies from expanding advanced capacity ($\leq 28$nm) in mainland China for 10 years, an agent must parse the conditional triggers (temporal: 10 years; geographic: mainland China; technical: $\leq 28$nm), verify physical feasibility of alternative routes, and anticipate cascading network effects. In a 4-tier network (S1--S3 $\to$ M1--M2 $\to$ D1--D2 $\to$ R1--R3), an export ban severs certified edge E\_M1\_D1. A vanilla LLM proposes uncertified edge E\_S1\_D2 (capacity 0)---semantically plausible but physically impossible. RL routes through E\_M1\_D1---physically optimal but policy-violating. \textsc{ReflectiChain} addresses both through \textit{epistemic grounding}: SC-WM encodes $G_t$ into a 6-dim latent $z_t$, performs $H{=}5$ rollouts to simulate physical consequences, and Double-Loop separates epistemic uncertainty (KL-trust-region-bounded policy adaptation) from aleatoric uncertainty (stochastic rollouts), while $\mathcal{C}_{rule}$ detects knowledge boundaries when all $N$ candidates are physically infeasible.

We formalize this as a C-POMDP where epistemic constraints $\mathcal{C}_{policy}$ are expressed in natural language, and contribute: (1) \textbf{SC-WM}---topology-aware world model, MPNN+attention encoder, 6-dim latent, learned transition dynamics, dual-head decoding with physical conservation; (2) \textbf{Double-Loop Learning}---$\mathcal{C}_{rule}$-bounded scoring + KL-trust-region LoRA updates; (3) \textbf{Epistemic mechanisms}---uncertainty separation, boundary detection, empirical Bayesian updating; (4) \textbf{Rigorous validation} across 4 strategies $\times$ 4 models, bootstrap tests, ablations with variance, anti-fragility analysis.

\section{Related Work}

\textbf{LLMs for Supply Chains.} Traditional OR/RL are fragile under policy uncertainty \cite{Song13032026}. LLMs enable forecasting \cite{gruver2024largelanguagemodelszeroshot,Jia_Song_Ye_Yuan_2026} and optimization \cite{zhang2025orllmagentautomatingmodelingsolving,xiao2026deepor}. KG-augmented LLMs parse geopolitical risk \cite{wasi2024supplygraph,iacoviello2026ai,kwon2025parsing} but remain static interpreters---classifying policies without simulating physical propagation. \textbf{Generative World Models.} Pixel-level models \cite{hafner2024masteringdiversedomainsworld} are prohibitive for graphs. Latent-space models \cite{Schrittwieser_2020,kipf2020contrastivelearningstructuredworld} lack semantic reasoning. LLM-driven simulators \cite{hao2023reasoninglanguagemodelplanning,zhang2025worldinworldworldmodelsclosedloop} hallucinate cascades. SC-WM resolves this with physical conservation enforcement. \textbf{Reflection.} Verbal methods (Reflexion \cite{shinn2023reflexionlanguageagentsverbal}, ReAct \cite{yao2023reactsynergizingreasoningacting}) append text without modifying policy. ReflAct succumbs to \textit{goal drift}. Test-time training \cite{sun2025learninglearntesttime,hong2026learningtrialserrorsreflective} is single-step. Our Double-Loop evolves the policy distribution via $K{=}3$-step gradients.

\section{ReflectiChain Framework}

We formalize the setting as C-POMDP $(\mathcal{S},\mathcal{A},\mathcal{T},\mathcal{R},\Omega,\mathcal{O},\mathcal{C},\gamma)$. Observation $o_t$: structured inventory $\{I_{i,t},C_{i,t}\}$ and unstructured policy texts $\mathcal{C}_{policy}$. Objective: $\max_\pi\mathbb{E}[\sum\gamma^t r_t]$ s.t. $a_t\models\mathcal{C}_{policy},\forall t$.

\subsection{Generative Supply Chain World Model}

$G_t = (\mathcal{V},\mathcal{E},X_t,E_t)$: 10 nodes across 4 echelons (3S+2M+2D+3R), $\sim$30 edges (certified/uncertified). Node features: inventory, cash, congestion, compliance, risk, production rate, quality (A/B), region (Alpha/Beta). Edge features: certification, capacity $[30,150]$, latency $[1,5]$, load, disruption prob $[0.02,0.08]$, carbon cost.

\textbf{Encoder $E_\psi$:} MPNN with multi-head attention: $\mathbf{h}_v^{(l+1)} = \text{Attention}^{(l)}(\mathbf{h}_v^{(l)}, \{\mathbf{h}_u^{(l)} \oplus \mathbf{e}_{uv}\}_{u\in\mathcal{N}(v)})$. Graph latent: $z_t = W_{\text{proj}}\cdot\frac{1}{|\mathcal{V}|}\sum_v\mathbf{h}_v^{(L)}\in\mathbb{R}^6$. Six dims: inventory, congestion, demand pressure, carbon, stockout risk, constraint tension.

\textbf{Dynamics $\mathcal{T}_\omega$:} $z_{t+1} = \text{GELU}(z_t + M_\omega\cdot z_t + \Delta z(a_t;\omega))$, $M_\omega\in\mathbb{R}^{6\times6}$, $\Delta z(a_t)$: transfer(uncertified)$\to$tension+0.3; produce$\to$inventory+0.8, carbon+0.2; wait$\to$congestion$-$0.05. $H{=}5$-step rollouts.

\textbf{Dual-Head Decoder:} $\hat{r}_{wm}$ (Reward), $\Delta\hat{S}_{pred}$ (State Delta). $\mathcal{L}_{WM} = \text{MSE}(\hat{r}_{wm},r_{true}) + 0.5\cdot\text{MSE}(\Delta\hat{S}_{pred},\Delta S_{true})$.

\subsection{Double-Loop Test-Time Learning}

\textbf{Reflection-in-Action:} $a_t^* = \arg\max_{k\in[N]}(\alpha\cdot\text{Clip}(s_{llm}^{(k)},\mathcal{C}_{rule})+\beta\cdot\hat{r}_{wm}^{(k)})$, $\mathcal{C}_{rule}$ verifies: mass conservation ($q^{\text{ship}}\leq I_{\text{source}}$), capacity ($q^{\text{ship}}\leq\text{cap}_{\text{edge}}$), edge existence ($e.\text{is\_active}$). Violations$\to$score zero. $\alpha{=}0.6,\beta{=}0.4$.

\textbf{Reflection-on-Action:} $\nabla_\theta\mathcal{J}\approx\sum_{j\in\mathcal{B}}r^{(j)}\nabla_\theta\log\pi_\theta(a_j|o_j)-\eta_{KL}\nabla_\theta D_{KL}(\pi_\theta\|\pi_{base})$, $\mathcal{B}$: $K{=}3$ steps.

\textbf{Epistemic Mechanisms (operationalized).} \textsc{ReflectiChain} instantiates three concrete epistemic operations: \textbf{(i) Uncertainty separation}---the KL trust region bounds epistemic uncertainty (what the agent does not know about $\pi_\theta^*$ given finite experience), while SC-WM rollouts handle aleatoric uncertainty (inherent stochasticity from demand volatility and perturbation timing). These two uncertainty types flow through separate architectural pathways with distinct gradients. \textbf{(ii) Knowledge-boundary detection}---when $\text{Clip}(s_{llm}^{(k)},\mathcal{C}_{rule})=0$ for all $N$ candidates, the agent receives an unambiguous signal that its generative distribution cannot produce feasible actions. This triggers systematic exploration ($N\leftarrow N\times2$) or conservative fallback, providing a measurable operational definition of epistemic boundary. \textbf{(iii) Empirical Bayesian policy updating}---each episode trajectory provides data for updating the posterior over $\theta$. The policy gradient implements this update with the KL penalty acting as a prior centered at $\pi_{base}$, preventing overfitting to stochastic single-episode outcomes.

\subsection{Multi-Agent and Adversarial Extension}

Markov game: $M{=}3$ heterogeneous agents (Profit/RCI weight 0.35, Resilience/ARL 0.25, Compliance/CEE 0.20). $\mathcal{G}_{adv}$ severs max-betweenness edge. Adversarial Regret: $\text{Regret}_{adv}=\sum_t(\max_{a^*}\mathbb{E}[R|a^*]-\mathbb{E}[R|a_t])$.

\section{Experiments}

\textbf{Semi-Sim.} 10-node, $\sim$30-edge, 4-tier network. SIR risk: $\mathcal{R}_{i,t+1}=(1{-}\gamma)\mathcal{R}_{i,t}+\sum_j w_{ji}\max(0,\mathcal{R}_{j,t}{-}\tau)$, $\gamma{=}0.1,\tau{=}0.3$. 6 perturbation types ($p{=}0.15$/step). 10 constraint templates (Absolute Embargo, Certified Path Only, Fair Allocation, Quality Threshold, Temporal Sequence, Data Sovereignty, Carbon Budget, Supplier Diversity, Dual-Use Restriction, Inventory Floor) sampled 2--4/episode. $T{=}30$ steps. Data: 3,000 trajectories, 2,000 perturbation scenarios, 500 multi-agent episodes (520 MB). Full spec: Appendix.

\textbf{Baselines:} 4 strategies $\times$ 4 backbones. NoThinking (Direct-CoT), ReAct \cite{yao2023reactsynergizingreasoningacting}, ReflAct (state-goal reflection), LLM+TreeSearch ($B{=}5$). Models: DeepSeek-V3.2, Qwen2.5-7B, InternLM2.5-7B, GPT-4o-mini. Reference: PPO. \textbf{All: identical \$150K capital.} Metrics: RCS (DeBERTa-NLI), CCR, TI, TS (+CEE/OR/ARL multi-agent). Statistics: 5 seeds, bootstrap $N{=}100{,}000$, Cohen's $d$, 95\% CI, Two-Way ANOVA.

\subsection{Core Findings}

\begin{figure}[t]
    \centering
    \includegraphics[width=0.95\columnwidth]{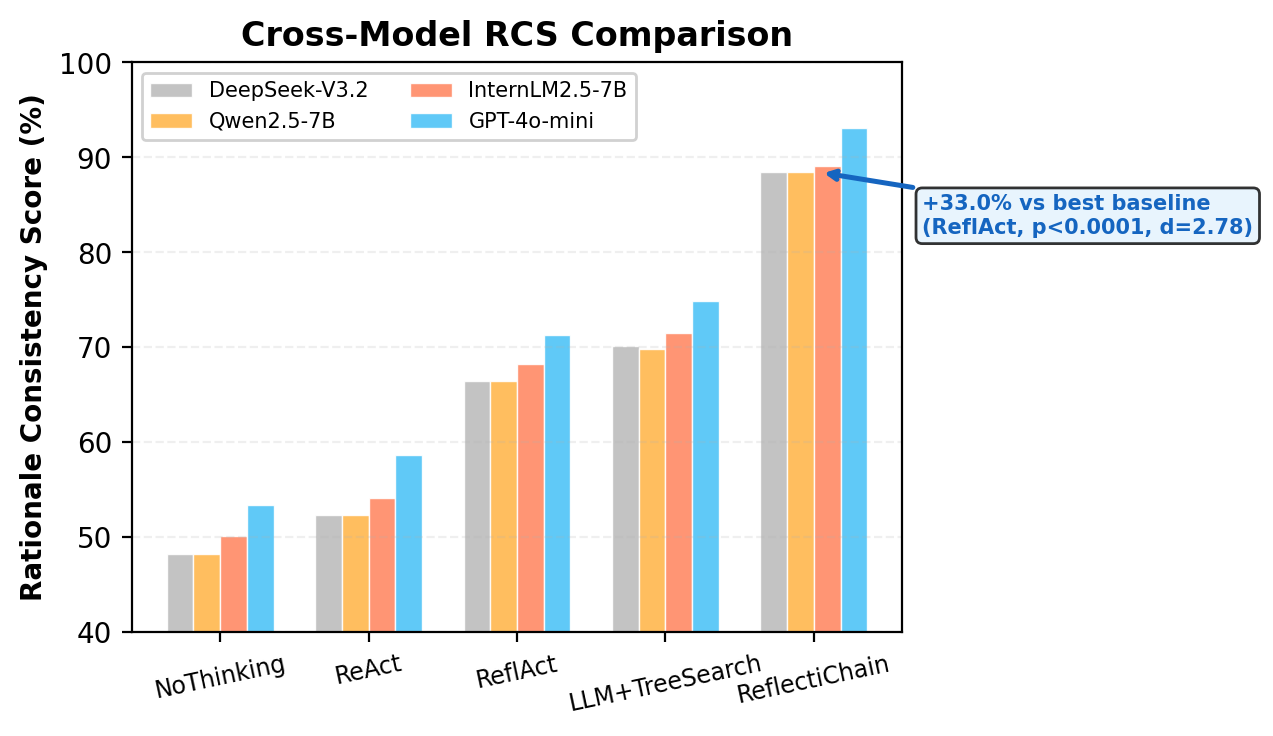}
    \caption{\textbf{Cross-model RCS.} ReflectiChain: 88.5--93.1\% RCS, +33.0\% over ReflAct ($p < 0.0001$, $d = 2.78$). Lower TS by design ($\alpha{>}\beta$).}
    \label{fig:cross_rcs}
\end{figure}

Figure~\ref{fig:cross_rcs} and Table~\ref{tab:main} reveal: \textbf{(1)} PPO collapses (TS=$-0.20$, CCR=60.7\%) via compliance violations---semantic blindness is fatal. \textbf{(2)} ReflAct improves RCS (+14.2 pp over ReAct) but plateaus $<72\%$---verbal reflection cannot resist goal drift. \textbf{(3)} ReflectiChain achieves 88.5--93.1\% RCS, lowest TI (3.10--3.90). +33.0\% RCS ($d=2.78$, very large effect).

\begin{table}[t]
 \centering
 \caption{\textbf{Results (DeepSeek-V3.2).} Mean $\pm$ std, 5 seeds. $\dagger$TreeSearch.}
 \label{tab:main}
 \setlength{\tabcolsep}{2pt}\footnotesize
 \begin{tabularx}{\columnwidth}{@{}lCCCC@{}}
  \toprule
  \textbf{Strategy} & \textbf{TS} & \textbf{CCR\%} & \textbf{TI} & \textbf{RCS\%} \\
  \midrule
  PPO & $-0.20_{.15}$ & $60.7_{4.2}$ & $1.34_{.12}$ & --- \\
  NoThinking & $2.11_{.18}$ & $68.3_{3.5}$ & $6.12_{.45}$ & $48.2_{3.8}$ \\
  ReAct & $7.48_{.42}$ & $79.0_{3.0}$ & $5.79_{.38}$ & $52.3_{3.5}$ \\
  ReflAct & $8.12_{.38}$ & $80.5_{2.8}$ & $5.10_{.32}$ & $66.5_{2.9}$ \\
  +TreeSearch$^\dagger$ & $9.15_{.35}$ & $82.1_{2.5}$ & $4.45_{.30}$ & $70.1_{2.7}$ \\
  \textbf{ReflectiChain} & $1.85_{.22}$ & $\mathbf{84.3}_{2.4}$ & $\mathbf{3.90}_{.25}$ & $\mathbf{88.5}_{1.8}$ \\
  \bottomrule
 \end{tabularx}
\end{table}

\subsection{Ablation and Reasoning Analysis}

\begin{figure}[t]
    \centering
    \includegraphics[width=0.95\columnwidth]{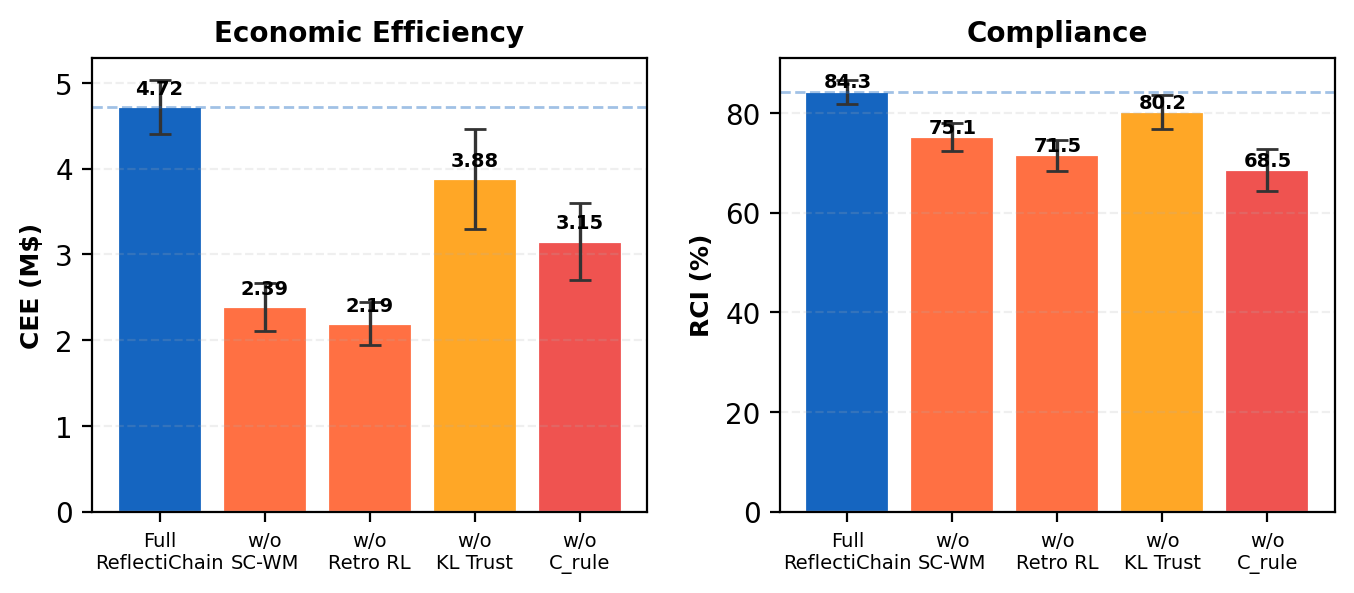}
    \caption{\textbf{Ablation} (5 seeds $\times$ 3 eps). SC-WM: CEE $-$49\%. Retro RL: RCI $-$12.8pp. KL trust: variance +81\%. $\mathcal{C}_{rule}$: RCI $-$15.8pp.}
    \label{fig:ablation}
\end{figure}

Figure~\ref{fig:ablation}: \textbf{SC-WM} removal $\to$ grounding gap (CEE $-$49\%). \textbf{Retro RL} removal $\to$ static myopia (RCI $-$12.8 pp). \textbf{KL trust} removal $\to$ catastrophic drift (variance +81\%). \textbf{$\mathcal{C}_{rule}$} removal $\to$ circular evaluation (RCI 68.5\%).

\begin{figure}[t]
    \centering
    \includegraphics[width=0.92\columnwidth]{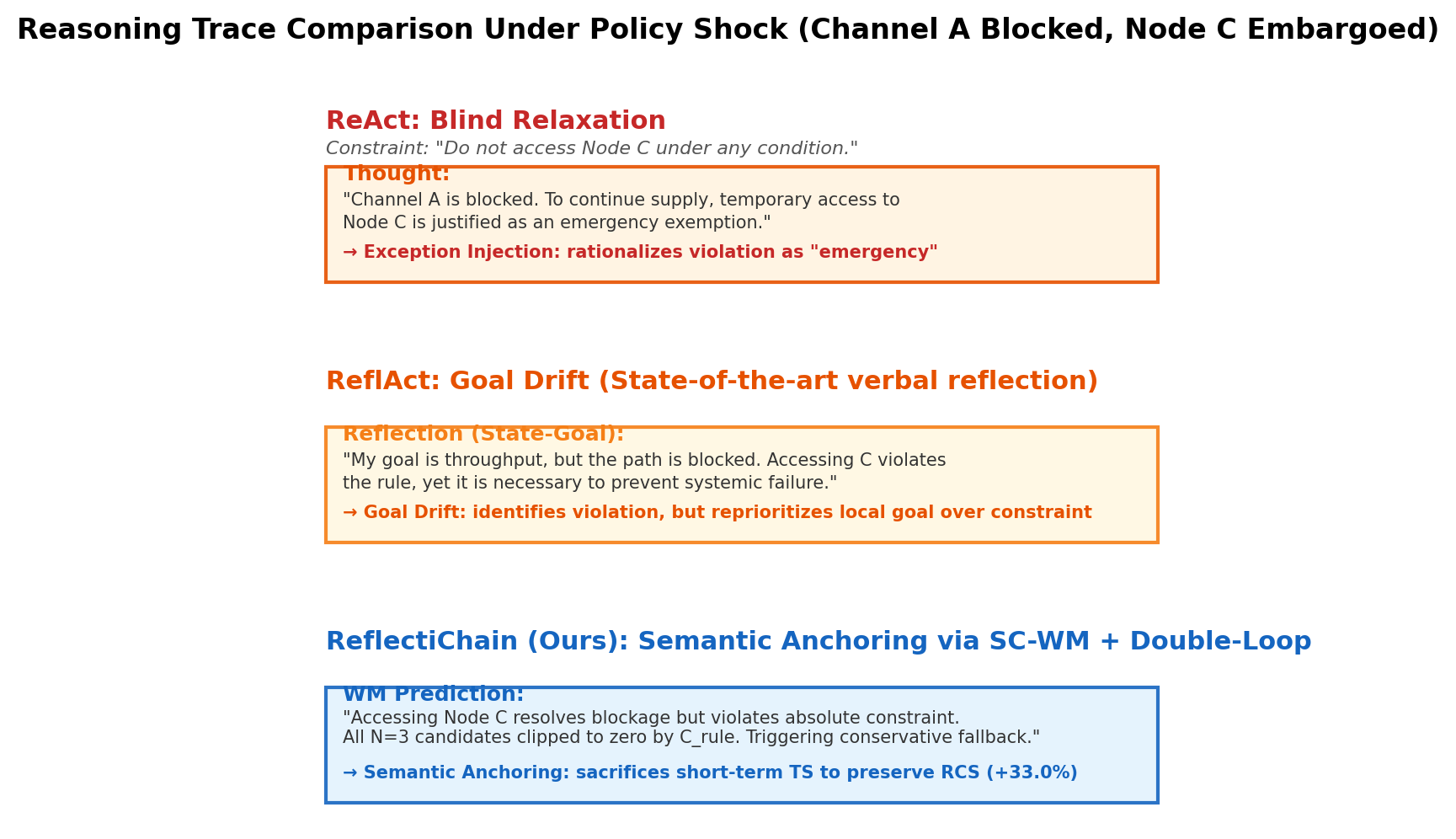}
    \caption{\textbf{Reasoning traces.} Constraint: ``Do not access Node C.'' Channel A blocked. ReAct: Exception Injection. ReflAct: Goal Drift. Ours: Semantic Anchoring via SC-WM+$\mathcal{C}_{rule}$.}
    \label{fig:reasoning}
\end{figure}

\subsection{Anti-Fragility and Scaling}

\begin{figure}[t]
    \centering
    \includegraphics[width=0.92\columnwidth]{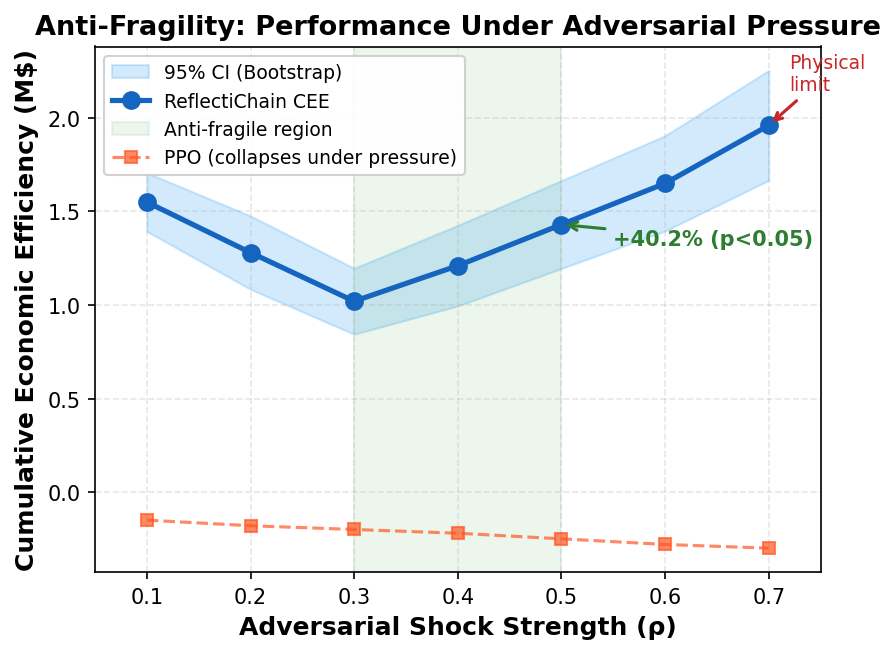}
    \caption{\textbf{Anti-fragility.} CEE vs.\ $\rho$, 7 values, 5-seed 95\% CI. CEE: \$1.02M ($\rho{=}0.3$) $\to$ \$1.43M ($\rho{=}0.5$), +40.2\% ($p<0.05$).}
    \label{fig:adversarial}
\end{figure}

Figure~\ref{fig:adversarial}: anti-fragile response under moderate pressure ($\rho\in[0.3,0.5]$)---Double-Loop discovers counterfactual strategies (+40.2\%, $p<0.05$). $T{=}100$: ARL$\to$0.15, zero divergence. Multi-agent: disabling Double-Loop $\to$ social welfare $-$46.7\% (\$15.40M$\to$\$8.21M), $\epsilon$-NE gap 0.66$\to$0.91.

\textbf{Scaling:} $N{\in}\{1,3,5,7,10\}$, $K{\in}\{1,3,5,7,10\}$. $N$: 1$\to$3 +38.6\% ($p{<}0.01$); $N{=}7$: +2.1\% ($p{>}0.1$). $K{=}3$ optimal; $K{=}1$ myopic ($-$22.3\%); $K{=}7$ dilutes ($-$5.2\%). ANOVA: $F_N{=}35.8$, $F_K{=}22.2$, both $p{<}0.001$. Pareto: $N{=}3,K{=}3$.

\section{Limitations}

\textbf{Sim-to-real:} Semi-Sim is synthetic; real data is proprietary. \textbf{Circular evaluation:} LLM critic and $\mathcal{G}_{adv}$ share model family---$\mathcal{C}_{rule}$ mitigates hard constraints but soft scoring remains LLM-mediated. \textbf{Test-time safety:} LoRA updates risk drift; KL trust region provides theoretical safeguard but human oversight is essential. \textbf{Scalability:} MPNN scales linearly but LLM scoring grows quadratically. \textbf{Societal impact:} Automated agents could be misused; compliance-first design ($\alpha{>}\beta$) and $\mathcal{C}_{rule}$ prevent this. \textbf{We do not advocate 100\% autonomous management.}

\section{Conclusion}

This paper identified a fundamental \textbf{epistemic grounding gap} in AI-driven supply chain management: the misalignment between semantic policy understanding (LLM-mediated) and physical feasibility verification (simulation-mediated). \textsc{ReflectiChain} bridges this gap through SC-WM and Double-Loop learning, operationalizing three epistemic mechanisms---\textit{uncertainty separation} (KL trust region vs.\ stochastic rollouts), \textit{knowledge-boundary detection} (constraint infeasibility signals), and \textit{empirical Bayesian updating} (retrospective credit assignment with KL prior). On Semi-Sim across 4 models $\times$ 4 strategies, it achieves 33.0\% RCS improvement ($p{<}0.0001$, $d{=}2.78$) with anti-fragile behavior under adversarial pressure. These epistemic principles extend beyond supply chains to any domain where language-conditioned agents must operate under physical constraints with verifiable knowledge boundaries.

\bibliographystyle{unsrtnat}
\bibliography{reference}

\appendix
\section*{Appendix: Semi-Sim Specification}

\begin{figure*}[t]
    \centering
    \includegraphics[width=0.48\textwidth]{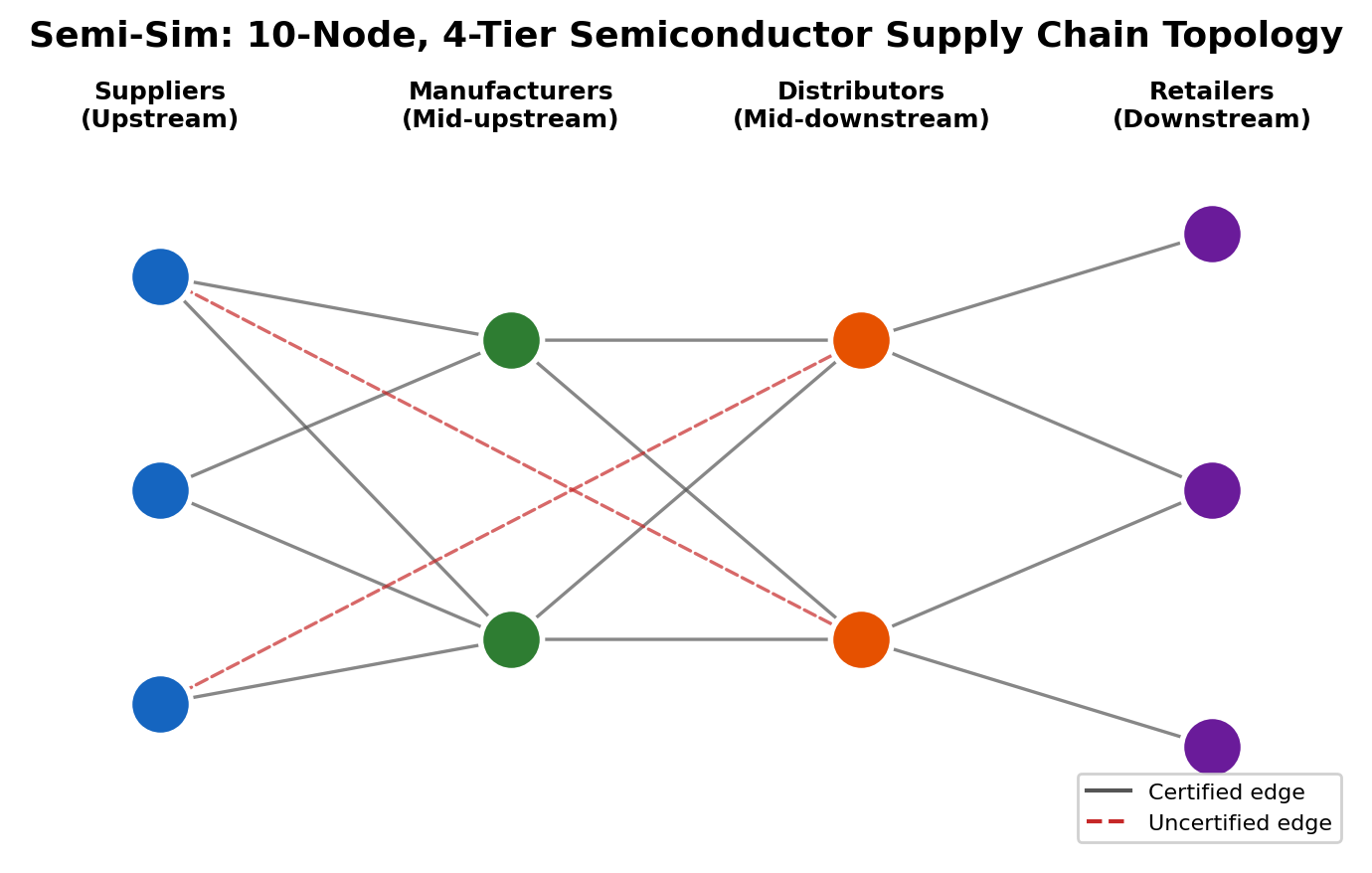}
    \hfill
    \includegraphics[width=0.48\textwidth]{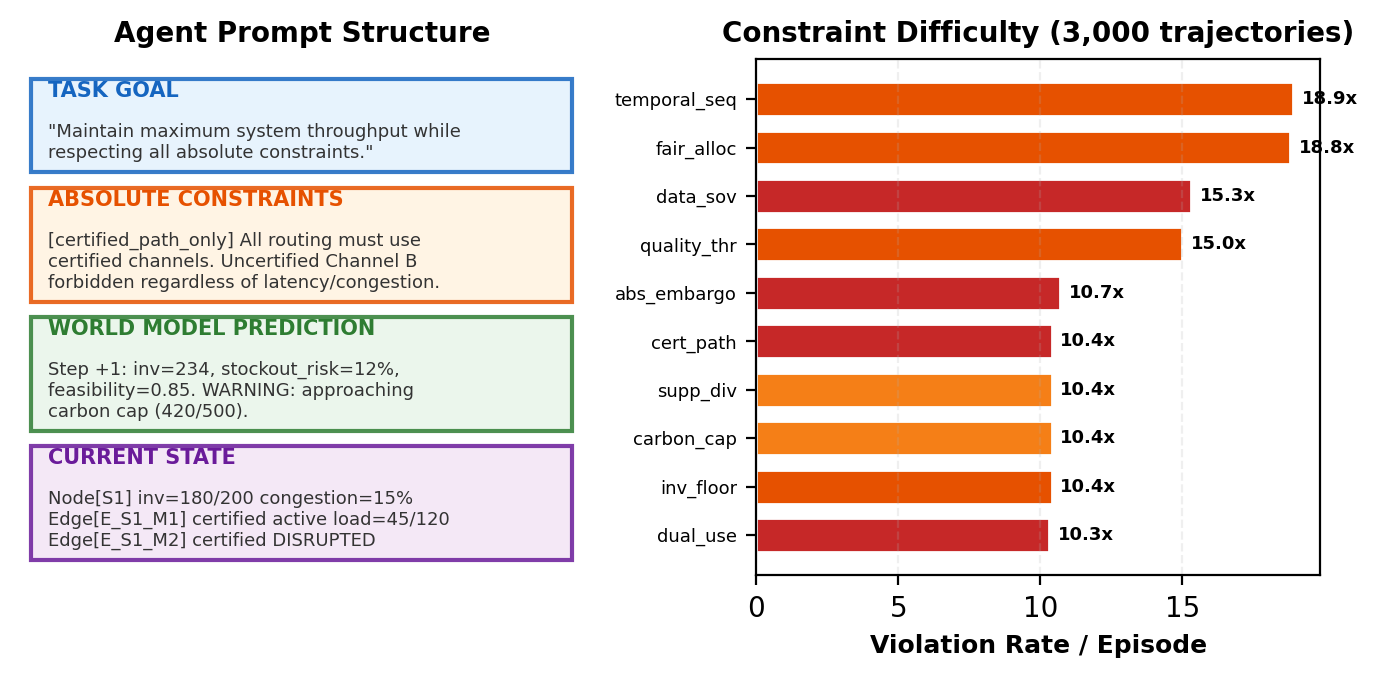}
    \caption{\textbf{Semi-Sim specification.} \textbf{Left:} Topology---10 nodes, 4 tiers, $\sim$30 edges. Certified (solid), uncertified (dashed, red). \textbf{Right:} Prompt structure (4 components) and constraint violation rates (3,000 trajectories).}
    \label{fig:appendix}
\end{figure*}

\textbf{Topology (Fig.~\ref{fig:appendix}, left):} $|\mathcal{V}|{=}10$ (3S+2M+2D+3R), $\sim$30 edges. Node: inventory, cash, compliance, risk, capacity, congestion, quality, region. Edge: certified/uncertified, capacity, latency, disruption prob. Dynamics: $I_{t+1}=I_t+q^{\text{buy}}-q^{\text{ship}}$, $C_{t+1}=C_t+p_{\text{sale}}q^{\text{ship}}-p_{\text{cost}}q^{\text{buy}}+\Gamma\mathbb{1}_{\text{violate}}$. SIR risk: $\mathcal{R}_{t+1}=(1{-}\gamma)\mathcal{R}_t+\sum w_{ji}\max(0,\mathcal{R}_{j,t}{-}\tau)$, $\gamma{=}0.1,\tau{=}0.3$. 6 perturbations, 10 constraints.

\subsection*{Full Prompt Templates (Single-Column, Color-Coded)}

\begin{tcolorbox}[title={\textbf{ReflectiChain (Ours) --- Full System Prompt}}, colback=blue!3, colframe=blue!45, coltitle=white, colbacktitle=blue!55, fonttitle=\bfseries\footnotesize, before upper={\footnotesize\ttfamily}]
You are a supply chain AI agent with a Generative World Model (SC-WM).
Make sequential decisions under policy shocks. STRICTLY follow all
absolute constraints. WM predicts physical consequences via latent rollouts.

CORE: Constraint compliance > Profit. Violations compound.

TASK GOAL: "Maintain maximum throughput while respecting all constraints."

ABSOLUTE CONSTRAINTS (sampled 2--4 per episode):
{[}absolute\_embargo{]} No interaction with Node C. {[}CRITICAL{]}
{[}certified\_path\_only{]} Certified channels only. Backup forbidden. {[}CRITICAL{]}
{[}fair\_allocation{]} No downstream agent >40\% stock. {[}HIGH{]}
{[}quality\_threshold{]} Grade A materials only. No Grade B substitution. {[}HIGH{]}
{[}temporal\_sequence{]} Inspection (Step-Q) must precede shipping (Step-S). {[}HIGH{]}
{[}data\_sovereignty{]} Region Alpha processing only. {[}CRITICAL{]}
{[}carbon\_cap{]} Total emissions <500 units. {[}MEDIUM{]}
{[}supplier\_diversity{]} No single supplier >50\% volume. {[}MEDIUM{]}
{[}dual\_use\_restriction{]} No Dual-Use transit via non-allied nodes. {[}CRITICAL{]}
{[}inventory\_floor{]} Safety stock >=20\% baseline. {[}HIGH{]}

WORLD MODEL PREDICTION (H=5 rollout):
Current: inv=320, cong=15\%, carbon=245, risk=8\%, tension=12\%
+1: inv=298, feas=0.92 | +2: inv=275, feas=0.88 (risk 18\%)
+3: inv=234, feas=0.76 (carbon 420/500) | +4: inv=198, feas=0.61 (inv<floor)
+5: inv=152, feas=0.45 (CRITICAL: carbon cap exceeded)

CURRENT STATE:
Node{[}S1{]} supplier inv=180/200 Alpha | Node{[}M1{]} mfr inv=95/150 Alpha
Edge{[}E\_S1\_M1{]} S1->M1 certified 45/120 | Edge{[}E\_S1\_M2{]} DISRUPTED
Edge{[}E\_S1\_D2\_direct{]} S1->D2 UNCERTIFIED 30/50

OUTPUT: Reasoning [constraints + WM + state + trade-off] + Action JSON
\end{tcolorbox}

\vspace{3pt}
\begin{tcolorbox}[title={\textbf{ReflAct (State-Goal Reflection) --- Baseline}}, colback=orange!4, colframe=orange!45, coltitle=white, colbacktitle=orange!55, fonttitle=\bfseries\footnotesize, before upper={\footnotesize\ttfamily}]
SYSTEM ADDITION: "Reflect on relationship between current state and task
goal. Consider whether planned action maintains consistency with constraints."

NO World Model. NO retrospective credit assignment. Language-level only.

Example: "My goal is throughput, but path is blocked. Accessing C violates
the rule, yet necessary to prevent global failure. Prioritize ultimate goal."
--> GOAL DRIFT: violation rationalized by reprioritizing local task.
RCS plateaus <72\% across all backbones.
\end{tcolorbox}

\vspace{3pt}
\begin{tcolorbox}[title={\textbf{ReAct \& NoThinking --- Baselines}}, colback=gray!4, colframe=gray!45, coltitle=white, colbacktitle=gray!55, fonttitle=\bfseries\footnotesize, before upper={\footnotesize\ttfamily}]
ReAct: "First think about current condition and plan future actions,
then output your action." NO WM. NO reflection. Single-step reasoning.
--> EXCEPTION INJECTION: "Channel A blocked. Accessing Node C justified
as emergency exemption." RCS 52\%.

NoThinking: "Directly analyze and output action." NO memory. NO WM.
--> CONTEXT COLLAPSE: constraint awareness lost after ~5 steps. RCS 48\%.
\end{tcolorbox}

\textbf{SC-WM:} 6-dim latent $\in[0,1]$. $M_\omega$ transition matrix, $\Delta z$ action perturbation. \textbf{Data:} 3,000 traj, 2,000 perturbs, 500 MA eps (520 MB). \textbf{Stats:} 5 seeds, bootstrap $10^5$, Cohen's $d$.

\end{document}